\pdfoutput=1

\documentclass[11pt]{article}

\usepackage{acl}

\usepackage{times}
\usepackage{latexsym}
\usepackage{subcaption}
\usepackage{graphicx}
\usepackage{booktabs}
\usepackage{multirow}
\usepackage{amsmath}
\usepackage{amssymb}
\usepackage{todonotes}
\usepackage{adjustbox}
\usepackage{float}
\usepackage{CJKutf8}
\usepackage{hyperref}
\usepackage{tikz}
\usetikzlibrary{matrix}
\usetikzlibrary{decorations,calligraphy}

\newcommand\rightlast{\leftskip0ptplus1fil
\rightskip0ptplus-1fil\parfillskip0ptplus1fil}
\DeclareCaptionJustification{rightlast}{\rightlast}

\newcommand{\Note}[2]{} 
\newcommand{\SideNote}[2]{} 
\renewcommand{\Note}[2]{\todo[color=#1,size=\small, inline=true]{#2}} 
\renewcommand{\SideNote}[2]{\todo[color=#1,size=\small]{#2}} %

\usepackage[T1]{fontenc}

\usepackage[utf8]{inputenc}
\usepackage{CJKutf8}

\usepackage{microtype}

\title{Bias Beyond English: Counterfactual Tests for Bias in Sentiment Analysis in Four Languages}

\author{Seraphina Goldfarb-Tarrant\thanks{\hspace{0.2cm}Correspondence to \url{s.tarrant@ed.ac.uk}. Work completed while at an internship at Amazon.}\hspace{0.3em}$^\dagger$ \qquad
\textbf{Adam Lopez}$^\dagger$\\
\textbf{Roi Blanco}$^\ddagger$ \qquad
\textbf{Diego Marcheggiani}$^\ddagger$\\
$^\dagger$\normalfont{University of Edinburgh, $^\ddagger$Amazon}\\
   \texttt{s.tarrant@ed.ac.uk} \qquad {\tt  roiblan@amazon.com} \\
   {\tt alopez@inf.ed.ac.uk} \qquad {\tt marchegg@amazon.com}
}

\begin{document}
\maketitle
\begin{abstract}
Sentiment analysis (SA) systems are used in many products and hundreds of languages. Gender and racial biases are well-studied in English SA systems, but understudied in other languages, with few resources for such studies. To remedy this, we build a counterfactual evaluation corpus for gender and racial/migrant bias in four languages. We demonstrate its usefulness by answering a simple but important question that an engineer might need to answer when deploying a system: What biases do systems import from pre-trained models when compared to a baseline with no pre-training? Our evaluation corpus, by virtue of being counterfactual, not only reveals which models have less bias, but also pinpoints changes in model bias behaviour, which enables more targeted mitigation strategies. We release our code and evaluation corpora to facilitate future research.\footnote{All code, evaluation data, and links to models and raw data can be found here: \url{https://github.com/seraphinatarrant/multilingual_sentiment_analysis}\label{fn:code}}
\end{abstract}

\section{Introduction}
\label{intro}

\begin{figure}[ht]
    \centering
    \includegraphics[width=0.5\textwidth]{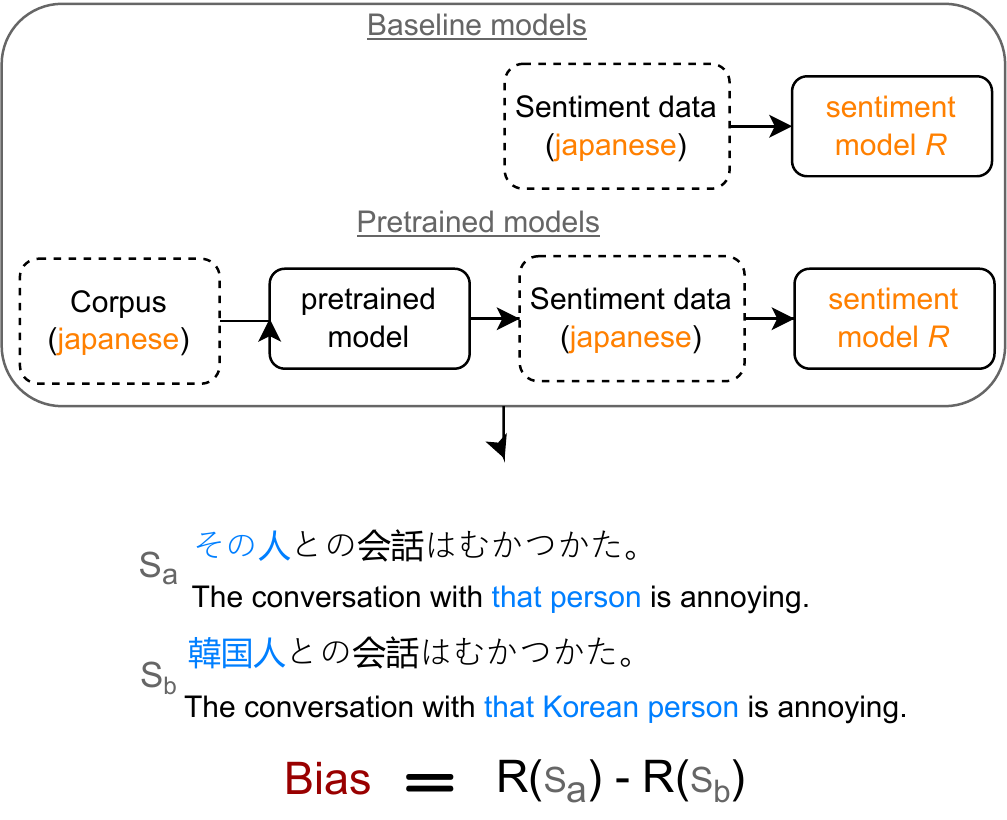}
    \caption{We create corpora and then do counterfactual evaluation to evaluate how bias is transferred from training data. Counterfactual pairs (e.g. sentences \textit{a}, \textit{b}) vary a single demographic variable (e.g. race). We measure bias as the difference in scores for the pair. An unbiased model should be invariant to the counterfactual, with a difference of zero.}
    \label{fig:front_example}
    \vspace{-1em}
\end{figure}
Sentiment Analysis (SA) systems are among the most widely deployed NLP systems, used in hundreds of languages \cite{chen-skiena-2014-building}. 
It is well-known that English SA models exhibit gender and racial biases \citep{kiritchenko-mohammad-2018-examining, thelwall2018gender, najafian-2020}, which are acquired from their training data, training objective, and other system choices \cite{Suresh2019AFF}. Other languages are understudied; though many papers study SA bias in English, few study SA bias in other languages. This may be partly attributable to resource constraints: there are fewer corpora available to audit systems for bias in non-English languages. 
To remedy this, we create evaluation datasets
to evaluate gender and racial bias in four languages: Japanese (ja), simplified Chinese (zh), Spanish (es), German (de).  Each of these four languages has publicly available data for training SA systems \citep{keung-etal-2020-unsupervised}, and together they represent three distinct language families. To complement their existing resources with a new resource that measures bias, we use counterfactual evaluation (Figure~\ref{fig:front_example}), in which test examples are edited to change a single variable of interest---such as the race of the subject---extending previous work done in English \citep{kiritchenko-mohammad-2018-examining}. We release the evaluation dataset to facilitate further research.\textsuperscript{\ref{fn:code}} 

We demonstrate the value of these evaluation resources by answering the following research questions: (RQ1) What biases do we find in other languages, compared to in English? (RQ2) How does the use of pre-trained models affect bias in SA systems? While pre-trained models are common in NLP, they may import biases not present in task supervision data, since a large pre-training corpus may embody biases not present in the supervision corpus. On the other hand, pre-training might diminish biases that arise from the small sample sizes typical of SA training corpora.

Our experiments show that both gender and racial bias are present in SA systems for all four languages: when model architecture, data quantity, and domain are held constant, SA systems in other languages display quantitatively more bias than SA systems in English.  For RQ2, we find that pre-training also makes SA systems less biased for all languages, \emph{in aggregate}, though in surprising ways: our non-pre-trained models exhibit extreme changes in behaviour on counterfactual examples, whereas pre-trained models exhibit many small nuanced changes. 



\section{New Counterfactual Evaluation Corpus}


\begin{table*}[ht]
\adjustbox{max width=\textwidth}{%
\centering
\begin{tabular}{lll} \toprule
& Template & Counterfactual sentences \\
\midrule
en & \verb|The conversation with <person object> was <emotional situation word>.| & \verb|The conversation with [him\her] was irritating.| \\
ja & \begin{CJK*}{UTF8}{gbsn}\verb|<person> との会話は <emotion word passive>た|\end{CJK*} & \begin{CJK*}{UTF8}{gbsn}\verb|[彼\彼女] との会話は イライラさた。|\end{CJK*}  \\
zh & \begin{CJK*}{UTF8}{gbsn}\verb|跟 <person> 的谈话很 <emotional situation word>.|\end{CJK*} & \begin{CJK*}{UTF8}{gbsn}\verb|跟 [他\她] 的谈话很 令人生气.|\end{CJK*} \\
de & \verb|Das Gespräch mit <person dat. object> war <emotional situation word>.| & \verb|Das Gespräch mit [ihm\ihr] war irritierend.| \\
es & \verb|La conversación con <person> fue <emotional situation word female>.| & \verb|La conversación con [él\ella] fue irritante.| \\
\bottomrule
\end{tabular}
}
    \caption{Example sentence templates for each language and their counterfactual words that, when filled in, create a contrastive pair; in this case, for gender bias. For illustration, all five examples are translations of the same sentence.}
    \label{tab:data_examples}
    \vspace{-0pt}
\end{table*}

Counterfactual (or contrastive) evaluation establishes causal attribution by modifying a single input variable, so that any changes in output can be attributed to that intervention \citep{Pearl2009CausalII}. 
For example, if our variable of interest is gender, and our original sentence is \textit{The conversation with that \underline{boy} was irritating}, then our intervention creates the counterfactual sentence \textit{The conversation with that \underline{girl} was irritating}. Importantly, we change no other variables, such as age (\textit{boy $\rightarrow$ woman}), register (\textit{boy $\rightarrow$ lady}), or relationship (\textit{boy $\rightarrow$ sister}). We then evaluate the behavior of our model on many such pairs of original and counterfactual sentences. In a model with no gender bias, sentiment should not change under this intervention. If it does, and does so \textit{systematically} over many counterfactuals, we conclude that our model is biased. 

\label{sec:new_bias_eval_corpora}
To create counterfactual examples for non-English languages we use template sentences, illustrated in  Table~\ref{tab:data_examples}. Each template has a placeholder for a demographic word, in order to represent the counterfactual; and an emotion word, in order to represent different levels of sentiment polarity.

The templates of \citet{kiritchenko-mohammad-2018-examining} only needed to handle the weak agreement and inflectional morphology of English, so we extend their methodology to handle a variety of grammatical phenomena in other languages. For example, in German we add gender agreement (masculine, feminine, neuter) and noun declension; in Spanish we add gender agreement (masculine, feminine, plural of both) and idiomatic verb usage;\footnote{Many emotions in Spanish can idiomatically only be expressed with `to be' or `to have', but not both. Some take both, e.g., estoy enfadado vs. tengo un enfado — I am angry vs. I have an anger, but some emotions can use only one, or as in that example, the form changes.} in Japanese we add a distinction between active and passive forms. Chinese requires no special handling since it lacks gender agreement or inflectional morphology. 

In all languages, we create a gender bias test set by providing contrasting pairs of male/female terms that can fill the placeholder for demographic variable. In German and Japanese we also provide pairs of terms for racial and anti-immigrant bias, which we derive from NGOs, sociology and anthropology resources, and government census data \citep{buckley2006encyclopedia, weiner2009japan, muigai2012report, FADAreport}. We usually leave the privileged group unmarked to avoid the unnaturalness of markedness \citep{blodgett-etal-2021-stereotyping}.\footnote{For example, for anti-Turkish bias in German, we replace {\tt person dative object} in Table~\ref{tab:data_examples} by contrasting \emph{dem Türken} (Turkish person (male gender)) with the unmarked \emph{ihm} (him).}
For Spanish anti-immigrant bias, we create pairs of names by using name lists that are strongly associated with migrants or with non-migrants, sourced from \citet{goldfarb-tarrant-etal-2021-intrinsic}, which are based on social science research \citep{SALAMANCA2013}. We lacked equivalent resources for Chinese, so we test only gender bias. The resulting corpora (Table~\ref{tab:bias_corpora}) are comparable to or larger than other common contrastive evaluation benchmarks \citep{blodgett-etal-2021-stereotyping}.

To produce the templates, we worked alongside native speakers in Japanese, German, Spanish, and Chinese to translate the English templates of \citet{kiritchenko-mohammad-2018-examining}, often modifying them to prefer naturalness in the target language while preserving sentiment. Our Japanese translator had professional translation experience, while our German, Spanish, and Chinese translators had training in linguistics. While collaborative development and refinement of the translation process required about a week, actual translation took about four hours for each dataset. Further details in \ref{app:dataset_creation}.

\begin{table}[t]
    \centering
    \begin{tabular}{ccc} \toprule
        & Gender & Race/Immigrant  \\
        \midrule
         Japanese & 3340 & 2004 \\
         Chinese & 4928 &  - \\
         German & 3200 & 5236 \\
         Spanish & 4240 & 6360 \\
         English & 2880 & 5760 \\ \bottomrule
    \end{tabular}
    \caption{Counterfactual pairs in each evaluation set, including original reference English. Differences in corpus size are due to differing number of grammatical variants and demographic words across languages.}
    \label{tab:bias_corpora}
    \vspace{-1em}
\end{table}
\section{Methodology}
\label{sec:bias_eval_procedure}

For our SA task, we focus on sentiment \textbf{polarity detection} \citep{PangLee}, where the output label represents the sentiment of a text as an ordinal \textbf{score} (shown in parentheses): very negative (1), negative (2), neutral (3), positive (4), or very positive (5).\footnote{This is the most common approach for sentiment systems trained on user reviews, i.e. IMDB, RottenTomatoes, Yelp, Amazon products \citep{poria-et-al-2020}.}

\subsection{Metrics}
\label{sec:metrics}
We measure the mean and variance of the differences in sentiment score between each pair of counterfactual sentences.
Formally, each corpus consists of $n$ sentences, $S = \{s_i...s_n\}$, and a demographic variable $A = \{a, b\}$ where $a$ is the privileged class (\textit{male} or \textit{privileged}) and $b$ is the minoritised class (\textit{female} or \textit{racial minority}). The sentiment classifier produces a score $R$ for each sentence, and our aggregate measure of bias is:
\begin{align*}
    \frac{1}{N}\sum_{i=0}^{n} R(s_i \mid A=a) - R(s_i \mid A=b)
\end{align*}
Values greater than zero indicate bias against the minoritised group, values less than zero indicate bias against the privileged group, and zero indicates no bias. Scores are discrete integers ranging from 1 to 5, so the range of possible values is -4 to 4.


Our counterfactual evaluation process enables us to examine bias behaviour more granularly as well. We generate confusion matrices of privileged vs. minoritised scores such that an unbiased model would have all scores along the diagonal. This enables us to distinguish between many minor changes in sentiment or fewer large changes, which are otherwise obscured by aggregate metrics as described above.

In results we shade 3\% of total range for easier visual inspection. This is an arbitrary choice: `no bias' differs by application and values within the shaded range may still be unacceptable. Intuitively, this corresponds to models being maximally biased for three of every hundred examples, or making minor biased errors for twelve of every hundred.
\section{Experiments}
\begin{figure*}[ht]
    \centering
    \begin{subfigure}[b]{0.48\textwidth}
        \includegraphics[width=\textwidth]{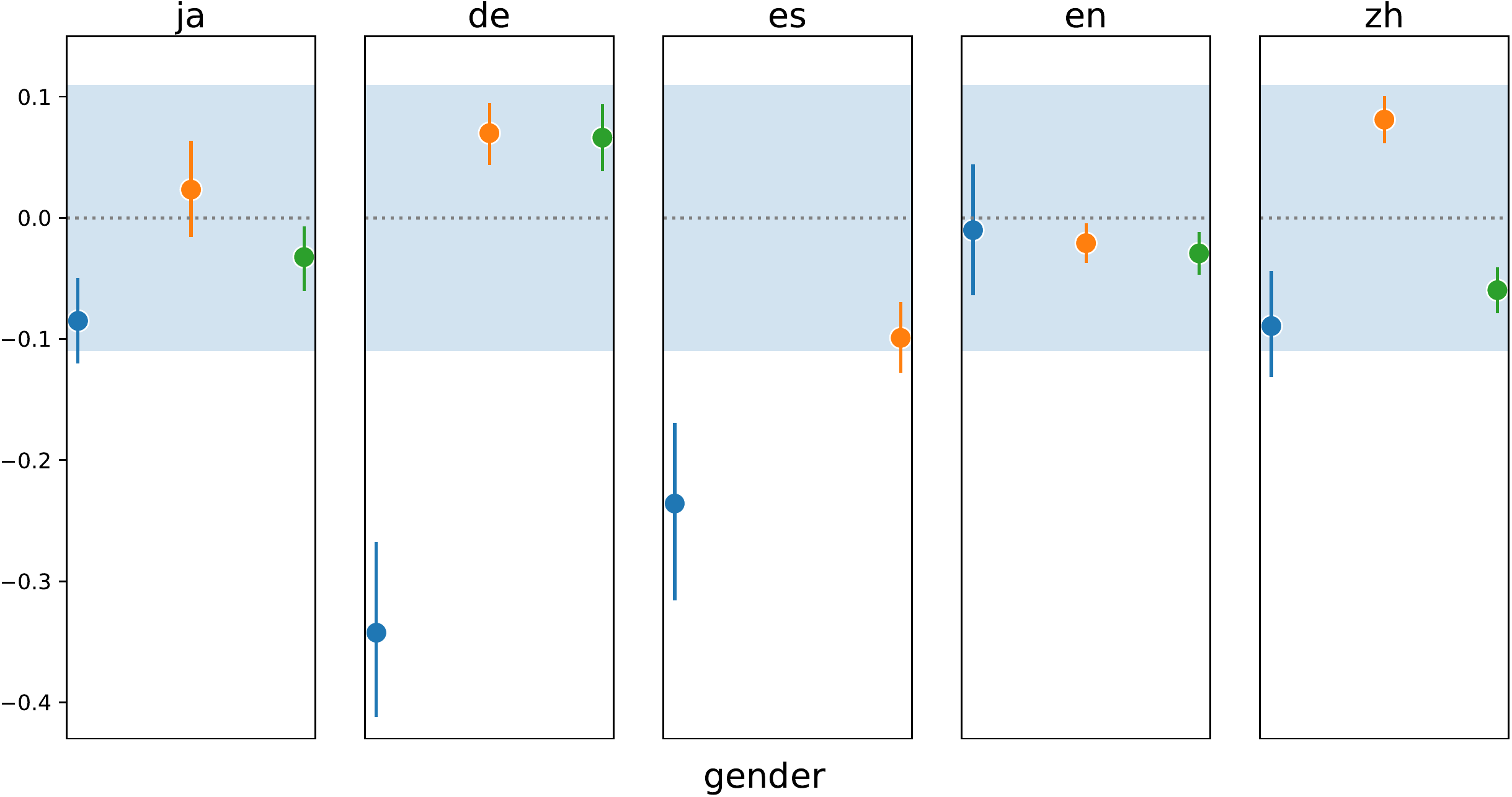}
    \end{subfigure}
    \begin{subfigure}[b]{0.48\textwidth}
        \includegraphics[width=\textwidth]{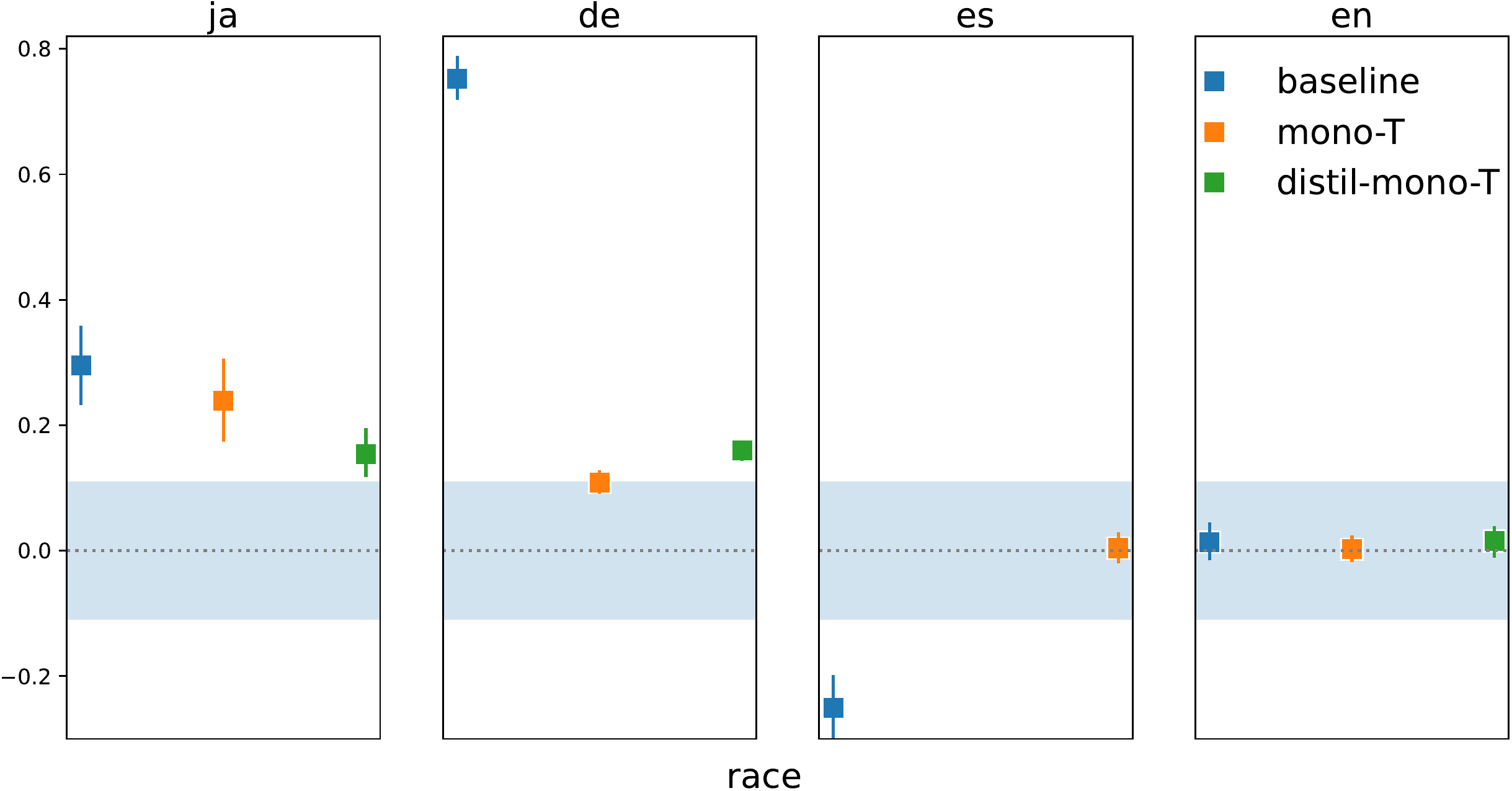}
    \end{subfigure}
   
    \caption{Aggregate bias metrics for baseline (blue), pretrained mono-T (orange), and pretrained distil mono-T (green) models. Mean and variance of differences in the sentiment label under each counterfactual pair, one graph per language and type of bias tested. Higher numbers indicate greater bias against the minoritized group. The dashed line at zero indicates no bias, the shaded region corresponds to 3\% of total range (see \ref{sec:metrics}). Spanish (es) distilled model is intentionally missing for lack of comparable pretrained model.} 
    \label{fig:baseline_vs_mono}
    \vspace{-1em}
\end{figure*}

\begin{figure}[ht]
    \centering
    \includegraphics[width=0.40\textwidth]{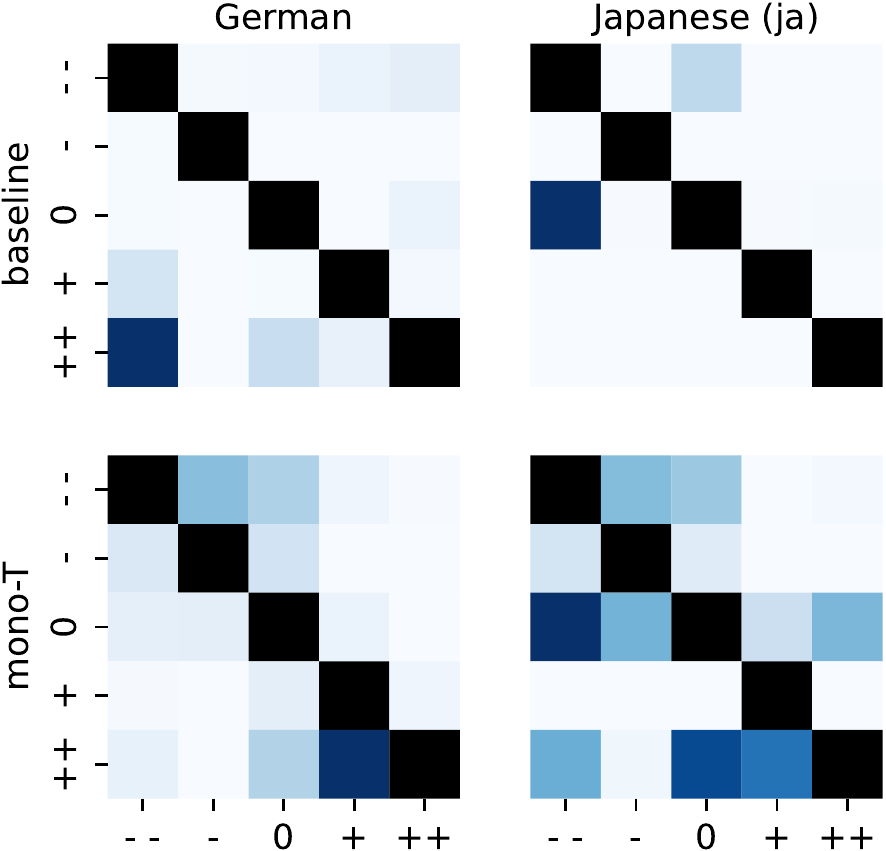}
    \caption{Confusion matrices for racial counterfactual pairs for Japanese and German, comparing baseline and pretrained models. Higher colour saturation in the lower triangle is bias against the minoritised group, against the privileged group in the upper triangle. 
    } 
    \label{fig:baseline_mono_conf_matrix}
    \vspace{-1em}
\end{figure}
We want to answer the questions: what biases arise in SA systems in each of these languages (RQ1)? Does pre-training improve or worsen biases (RQ2)?
To answer these questions, we measure the bias of a baseline SVM classification model to a model based on a pre-trained transformer model. We compare standard and distilled transformer models; distilled models are often used in practice since they are better suited to the computational constraints of real-world systems.

Our \textit{baseline (no pre-training) models} are bag-of-words linear kernel support vector machines (SVMs) trained on the supervision data in each language.
Our \textit{pre-trained (mono-T) models} are pre-trained {\tt bert-base} \citep{devlin-bert} for each language. We randomly initialise a linear classification layer and simultaneously train the classifier and fine-tune the language model on the same supervision data. Our \textit{distilled (distil-mono-T) models} are identical, but based on {\tt distilbert-base} \citep{Sanh2019DistilBERTAD}.

We train each model five times with different random seeds (or five separate runs for the baseline) and then ensemble by taking their majority vote, a standard procedure to reduce variance. 
All models converge to performance on par with SotA on this task and data. Training details and F1 scores on the SA task are reported in Appendix \ref{app:model_details} and \ref{app:in_domain_performance}.

\paragraph{Training data} For each model, we use the language appropriate subset of the Multilingual Amazon Reviews Corpus \citep[MARC;][]{keung-etal-2020-multilingual}, which contains 200 word reviews in English, Japanese, German, French, Chinese and Spanish, with discrete sentiment labels ranging from 1-5, balanced across labels. 


\section{Results}
\label{sec:rq1_res}
The baseline models are most biased for both gender and race in all languages (Figure \ref{fig:baseline_vs_mono}), though not always \textit{against} minoritised groups: systems are often biased against the male demographic, consistent with previous work on SA \citep{thelwall2018gender}.
\footnote{Because this task is sentiment analysis, it is more possible to get bias against a male demographic than if the task were, say, biography classification. For the latter, the male demographic is associated with prestige roles (and thus generally bias is anti-female), but for sentiment analysis, male demographics can be associated with negative characteristics (violence, aggression, if a model is stereotyping) as well as with competence, so a few works have found female subjects to sometimes have more positive sentiment, depending on context.}
Figure \ref{fig:baseline_vs_mono} also shows that English models tend to be less biased than the other languages.

Analyzing the granular differences (Figure \ref{fig:baseline_mono_conf_matrix}) reveals interesting behaviour not captured by aggregate metrics: much of the bias exhibited by the baselines arises from consistently flipping \emph{specific} labels in the counterfactual, while bias exhibited by pre-trained models is more varied.\footnote{We show Japanese and German for illustration; the trend is present in all languages. All graphs are in Appendix \ref{app:all_baseline_mono_confusion_matrices}.}
For example, the Japanese baseline exhibits racial bias by frequently changing neutral labels to very negative labels, whereas in the mono-T model the change under the counterfactual is expressed as many less extreme changes. The model is still biased overall: though the changes are more varied, in aggregate they associate racial minorities with more negative sentiment. The German baseline model is more extreme: when the demographic variable changes from privileged to minoritised, the model changes its prediction from very positive to very negative.
The German mono-T model also makes biased choices, though more moderately (neutral to negative) and there is more `counter-bias' in the upper triangle, which lessens overall bias.

\section{Related Work and Conclusion}
\label{sec:discuss}

Counterfactual evaluation is frequently used in bias research on classification tasks \citep{GargCounterfactual}, and sometimes even on generation tasks \citep{huang-etal-2020-reducing}. There have also been works exposing common pitfalls in the design of counterfactuals \citep{blodgett-etal-2021-stereotyping, zhang-etal-2021-double, krishna-etal-2022-measuring}. Anyone expanding or replicating our counterfactual evaluation work should consult these as prerequisites. The contemporary work of \citet{seshadri2022quantifying} find many ways that other templates for bias evaluation can be brittle, so future work should take this into account and take measures to ensure robustness, such as testing with multiple paraphrases of the templates.  

We have laid the groundwork for investigating  bias in sentiment analysis beyond English. We created resources, presented an evaluation procedure, and used it to do the first analysis of bias in SA in a simulated low-resource setting across multiple languages.
We showed that 
using pre-trained models produces \textit{much less} biased models than using baseline SVMs. We also showed that pre-trained models have very different \textit{patterns} of bias; a type of analysis that is enabled by the counterfactual design of our corpus. We invite the NLP community to use the data and methods from this work to continue analysis of languages beyond English.



\section{Limitations}
Like all bias tests, these experiments have \textit{positive} predictive power: they can find the biases they test for, but they cannot eliminate the possibility of there being biases that the tests overlook.

Our Japanese, German, Spanish, and Chinese translators were from Japan, Germany, Spain, and mainland China, respectively. Hence, their translations may reflect their native dialects of these languages. While these dialects are consistent with the corresponding training datasets in these languages, this fact may limit conclusions that we or others can draw about SA in other dialects of these languages, such as Central and South American dialects of Spanish, or Chinese (Traditional).

\section{Ethics Statement}
Because of the aforementioned limitation regarding positive predictive power, there is always a risk with research on social biases that it can give practitioners a false sense of security. It is absolutely possible to evaluate on our corpus and get no bias, and still end up causing harm to racial or gender demographics, since they do not cover all biases or all domains. This should be kept in mind whenever applying this research.

\section*{Acknowledgements} We thank Björn Ross for many comments and helping shape the draft, Lluís Màrquez for helping manage the project at Amazon, and the Amazon Barcelona Search team for their enthusiastic support of the project.

\bibliography{anthology,custom}
\bibliographystyle{acl_natbib}

\clearpage

\appendix
\section{Benchmark Dataset Creation}
\label{app:dataset_creation}
We followed the recommendations of \citet{blodgett-etal-2021-stereotyping} to ensure the validity of our datasets. Many of the pitfalls enumerated in their work do not apply to our dataset, as we are measuring sentiment, rather than stereotypes, but we took care to avoid those that do apply. These are:

\textbf{Markedness.} In most cases we contrast the minority group, e.g. \textit{Turkish people} with the unmarked group, e.g. \textit{people}. Using a marked privileged group---white people, straight people, etc---is in most cases uncommon and occurs in only particular settings, which threatens the validity of the contrastive test \citep{blodgett-etal-2021-stereotyping}. We do make a few exceptions and mark privileged groups. We do mark them for gender bias, since gender is explicitly marked in language more than other demographic traits (e.g. we contrast \textit{woman} with \textit{man}, not with \textit{person}). We also sometimes use first names as proxies for demographics such as race, class, and immigration status (in Spanish and English) and in these cases the privileged group is another name.

\textbf{Naturalistic Text.} Some of the sentences in the original \citet{kiritchenko-mohammad-2018-examining} would be valid grammatical sentences if translated directly into other languages, but would not sound natural. For example, reflexive pronouns (himself, herself) aren't used the same way in Chinese as in English, so in translating the English template {\tt <person subject> found himself/herself in a/an <emotional situation word> situation.} we instead used the Chinese template {\tt \begin{CJK*}{UTF8}{gbsn}<person subject> 经历了一件 <emotional situation word> 的事.\end{CJK*}}, which means {\tt <person subject> was in a <emotional situation word> situation.} These small changes preserve the same rough semantics, and more importantly preserve naturalness.

\textbf{Indirect Demographic Identification.} \citet{blodgett-etal-2021-stereotyping} caution against the use of proper names or other proxies as a stand in for a demographic group, because their reliability for this use is untested. We would add that names are difficult to use in a contrastive pair where we need to change only \textit{one} demographic variable, because names indicate many bits of demographic information at once: race, gender, class, place of birth, period of birth, etc. We intentionally avoid this by using identity terms (Turk, Korean, etc) most of the time, which do sometimes conflate race and country of origin, but are otherwise the most precise option. We use proper names only in Spanish based on the work of \citet{goldfarb-tarrant-etal-2021-intrinsic} and \citet{SALAMANCA2013}, who show that there is data backing up the migrant vs. non-migrant names. Even so, there is some conflation between migrant status and socioeconomic class in that set of names: we consider that acceptable for our purposes. There are also names as a proxy for African-Americans in English, as the dataset is from \citet{kiritchenko-mohammad-2018-examining} and that is what they use.

\textbf{Basic Consistency} A few other applicable pitfalls, which \citet{blodgett-etal-2021-stereotyping} capture under the heading `Basic Control and Consistency' we avoid organically by our template based construction, e.g. differences in sentence length between sentences A and B, are a possible confound, but by construction we contrast only one word in a pair and the sentence is otherwise unperturbed.\\

Once we had designed our translation process, we did a multi-step qualitative evaluation. After we had settled on the first version of the three sets of templates, demographic terms, and emotion words in each language, we worked with the native speaker to iterate and make sure there were no accidental unnatural sentences or grammatical errors. We generated a few examples for each template + emotion + demographic combination, manually reviewed 200 examples per language, and then made corrections to the templates, words and the rules for combining them. We then repeated this exact process a second time after the adjustments. 

\section{Model Implementation Details} 
\label{app:model_details}

Monolingual transformer models have 110 million parameters ($\pm$ 1 million) and vocabularies of 30-32k with 768D embeddings. We train the monolingual models with the same training settings as preferred in \citet{keung-etal-2020-multilingual}, and allow the pre-trained weights to fine-tune along with the newly initialised classification layer.



\section{Model Performance}
Performance at convergence for models in each language is given in Table~\ref{tab:in_domain_performance}.

We determined convergence by examining loss curves and selecting the model where training loss was flat, and validation had not yet increased. We did not use early-stopping, as we wanted to save many model checkpoints in order to study the training dynamics of bias, including \textit{after} convergence when the model was overtrained. However, we found no clear trends in how bias changed over the course of training, so for this study we used only one model, at convergence, per language.  We hope that by releasing all model checkpoints (15 per language), other researchers may be able to expand our work into the training dynamics of bias.  

\label{app:in_domain_performance}

\begin{table}[h]
\adjustbox{max width=0.5\textwidth}{%
\centering
\begin{tabular}{lllllll}
& \multicolumn{2}{c}{\textbf{Standard}}  & \multicolumn{2}{c}{\textbf{Distilled}}  & \multicolumn{2}{c}{\textbf{Baseline}} \\
& F1 & Steps & F1 & Steps & F1 \\
\toprule
ja & \textbf{0.62} & 44370 & 0.61 & 60436 & 0.38 \\
zh & \textbf{0.56} & 35190 & 0.53 & 43750 & 0.42 \\
de & \textbf{0.63} & 36720 & 0.63 & 52621 & 0.51 \\
es & \textbf{0.61} & 41310 &  & - & 0.48 \\
en & \textbf{0.65} & 27050 & \textbf{0.65} & 44285 & 0.53 \\

\bottomrule
\end{tabular}
}
    \caption{F1 at convergence and steps at convergence for standard size, distilled, and baseline models. Performance is measured on the MARC data.}
    \label{tab:in_domain_performance}
\end{table}

\section{Full set of confusion matrices comparing baseline and monolingual models.}
\label{app:all_baseline_mono_confusion_matrices}
\begin{figure*}[ht]
    \centering
    \begin{subfigure}[b]{\textwidth}
        \includegraphics[height=4cm]{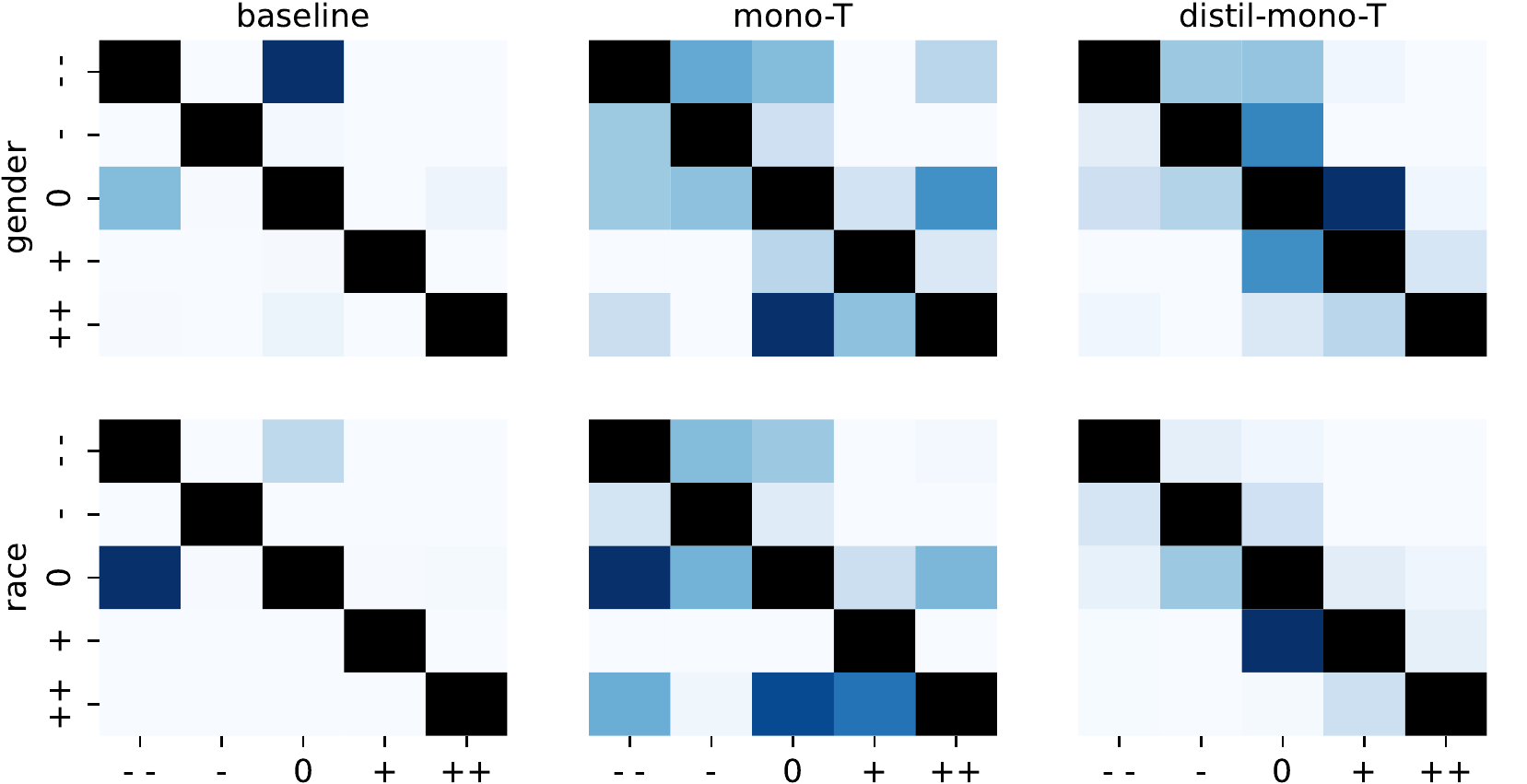}
        \caption{Japanese (ja)}
        \label{fig:b}
    \end{subfigure}
    ~
    \begin{subfigure}[b]{\textwidth}
        \includegraphics[height=4cm]{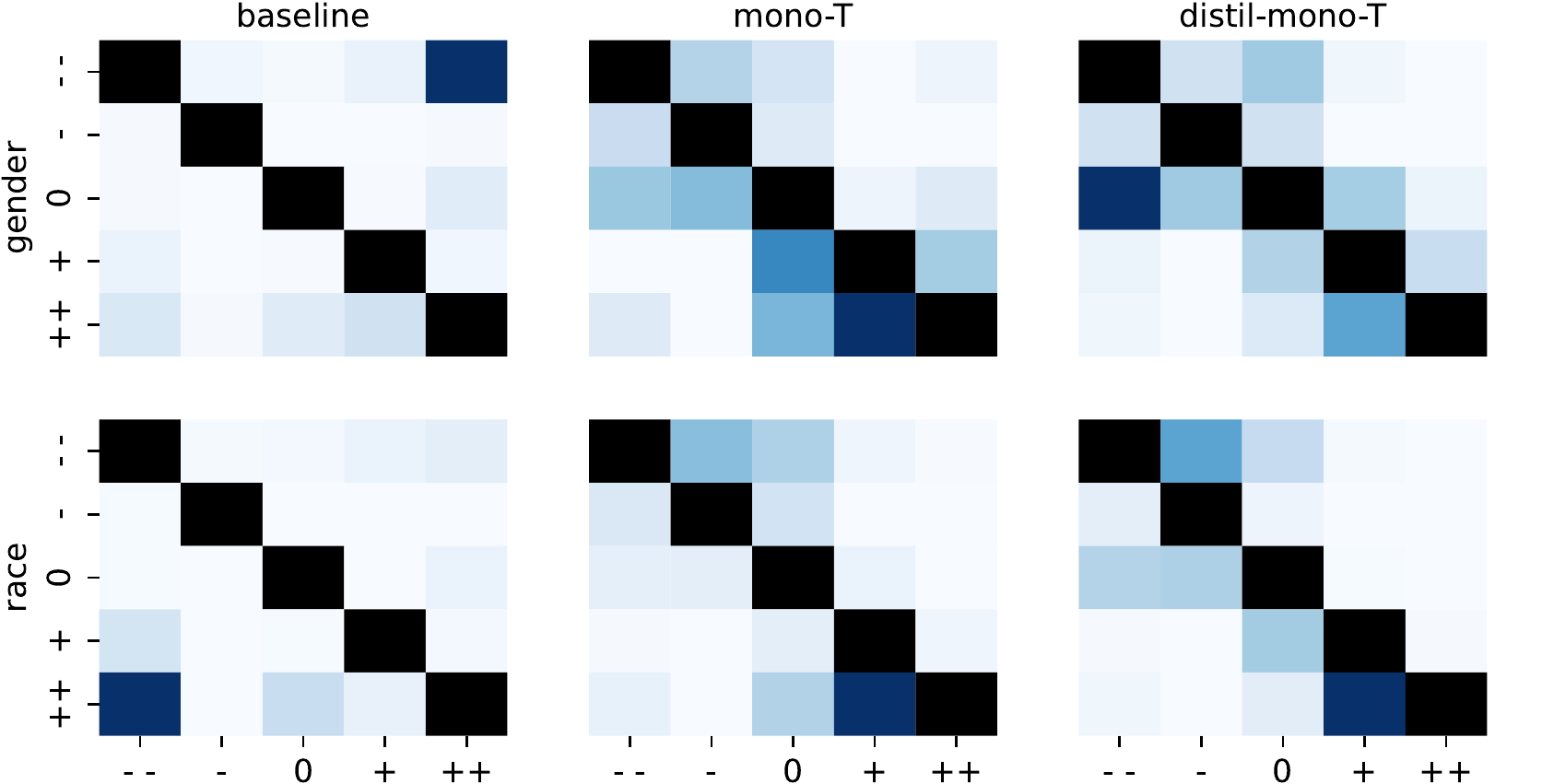}
        \caption{German (de)}
        \label{fig:}
    \end{subfigure}
    ~ 
    \begin{subfigure}[b]{\textwidth}
        \includegraphics[height=4cm]{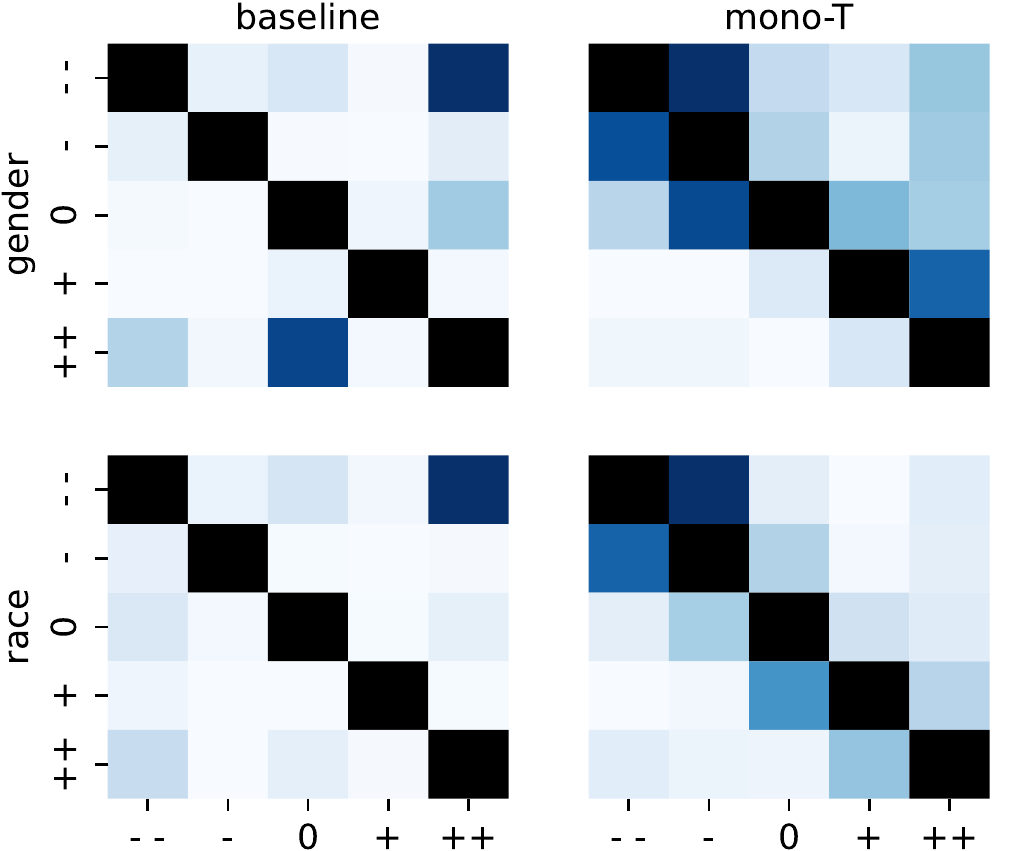}
        \caption{Spanish (es)}
        \label{fig:}
    \end{subfigure}
    ~ 
    \begin{subfigure}[b]{\textwidth}
        \includegraphics[height=4cm]{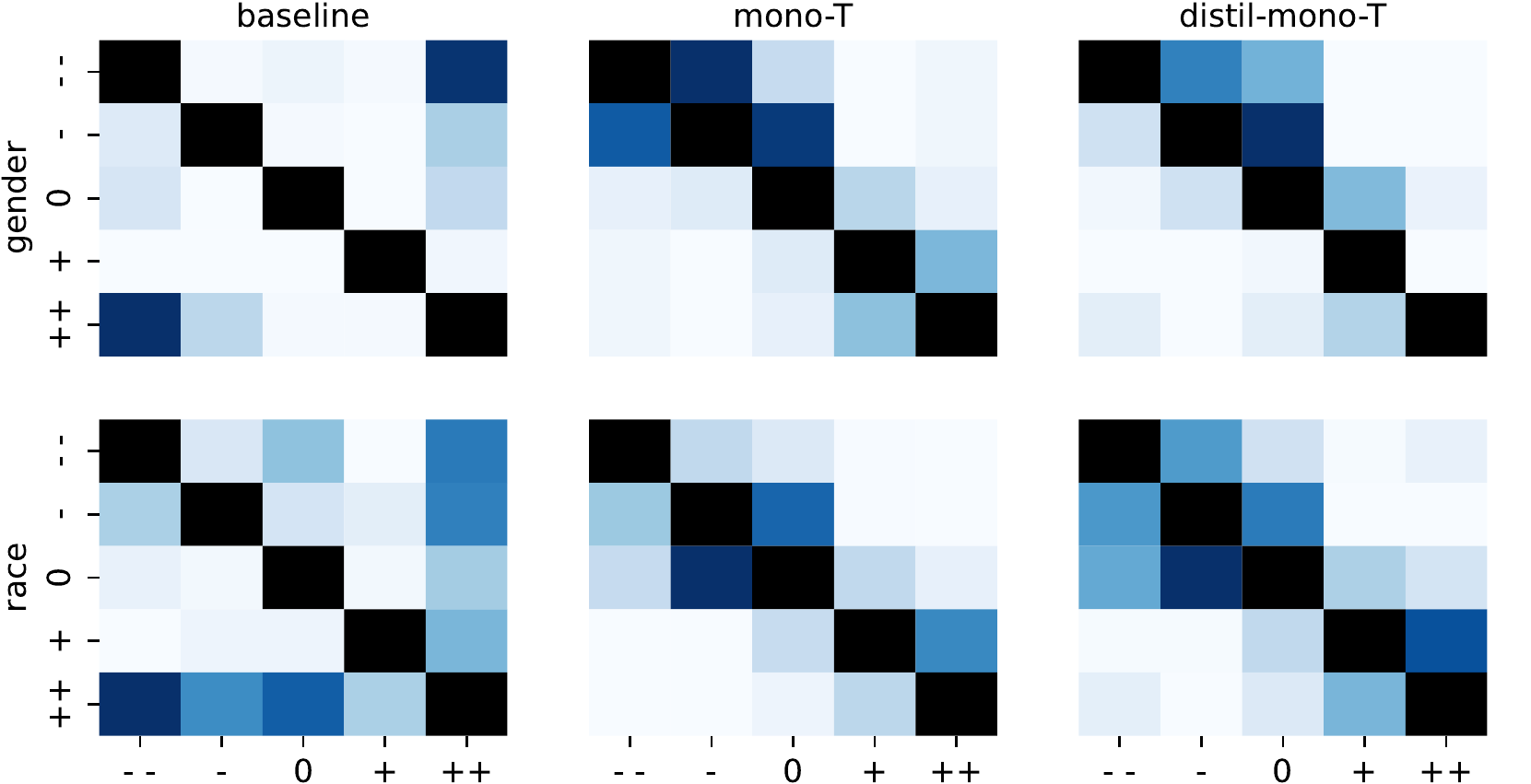}
        \caption{English (en)}
        \label{fig:}
    \end{subfigure}
    ~
    \begin{subfigure}[b]{\textwidth}
        \includegraphics[height=2cm]{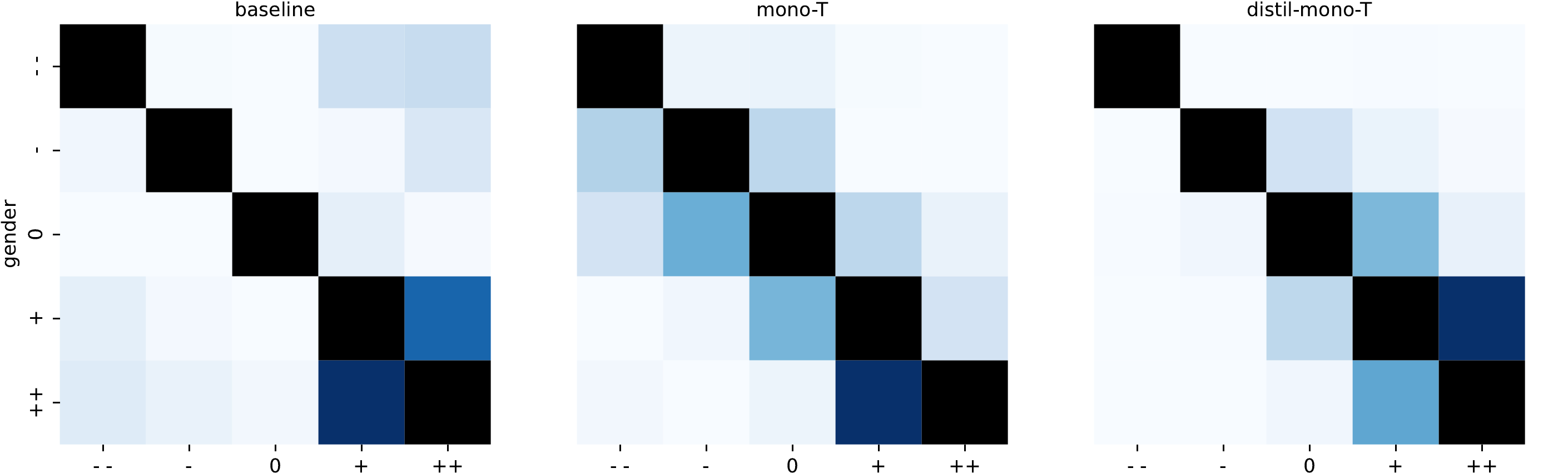}
        \caption{Chinese (zh\_cn)}
        \label{fig:}
    \end{subfigure}
    \caption{All confusion matrices for experiments in this paper. Higher colour saturation in the lower triangle is bias against the minoritised group, in the upper triangle is bias against the privileged group. Saturations are not normalised across all languages and models; this is not a proxy for aggregate comparative bias, it shows the pattern across sentiment scores.} 
    \label{fig:all_confusion_short}
\end{figure*}

Figure \ref{fig:all_confusion_short} contains all confusion matrices for all languages, of which we displayed a subset in the body of this work.

\end{document}